# MR-JEPA: A General Purpose Video Foundation Model for Cardiac MRI

Athira J. Jacob[1,2], Puneet Sharma[1], Dorin Comaniciu[1], and Daniel Rueckert[2,3]

[1] Digital Technology and Innovation, Siemens Healthineers, Princeton, NJ, USA
[2] Chair for AI in Healthcare and Medicine, Technical University of Munich (TUM) and TUM University Hospital, Munich, Germany
[3] Department of Computing, Imperial College London, UK

**Abstract.** Cardiac magnetic resonance imaging (CMR) produces rich sequential data such as temporal cine videos and spatial LGE/mapping stacks, yet most deep learning approaches process individual 2D slices, discarding this context. We present MR-JEPA, a self-supervised video foundation model for CMR that extends LeJEPA to 3D spatiotemporal inputs through tubelet tokenization, spatiotemporal masking augmentation, and initialization from a 2D CMR foundation model. Unlike prior CMR video models limited to cine data, MR-JEPA is pretrained on multi-sequence data (cine, LGE, mapping) from 10,505 patients across two centers without annotations. We evaluate the frozen encoder on six downstream tasks using a unified multi-view gated attention architecture: LV ejection fraction, RV ejection fraction, three myocardial strains (GLS, GCS, GRS), and four-class disease detection. MR-JEPA outperforms other compared methods on all five regression tasks, including both a domain-specific CMR model pretrained on more data with text supervision and a natural-video foundation model, achieving an LV EF MAE of 4.79% ($r$=0.764) and a GLS MAE of 1.87 ($r$=0.805), with 21–27% MAE reductions over baselines on strain tasks. For disease detection, MR-JEPA achieved a macro AUC of 0.868, remaining competitive with the domain-specific baseline despite using a fully self-supervised pretraining objective. These results demonstrate the potential of a unified video encoder for robust, multi-view utilization of diverse CMR sequences in clinical cardiac quantification and diagnosis.

**Keywords:** Cardiac MRI · Video Foundation Model · Strain Regression · Disease Detection.

## 1 Introduction

Cardiac magnetic resonance (CMR) provides complementary assessment of cardiac anatomy, function, and tissue characterization through multiple sequence types, including cine, late gadolinium enhancement (LGE), and parametric mapping. These sequences contain rich temporal and spatial information relevant to cardiac function, myocardial mechanics, and disease characterization [16].

Most deep learning methods for CMR operate on individual 2D slices [9], disregarding the sequential context inherent in cine and multi-slice acquisition. Recently, self-supervised video foundation models have demonstrated strong spatiotemporal representation learning from unlabeled data [4, 24]. However, adapting these to CMR remains challenging due to limited medical video data, heterogeneous imaging sequences, & the need to integrate information across multiple views and contrasts.

Prior work on CMR video foundation models has largely focused on cine imaging and often relies on auxiliary text supervision. In this work, we introduce MR-JEPA (Magnetic Resonance Joint-Embedding Predictive Architecture) (Fig. 1), a fully self-supervised video foundation model for CMR based on LeJEPA (Latent-Euclidean JEPA) [5]. We extend LeJEPA to spatiotemporal CMR inputs using tubelet embeddings and spatiotemporal masking, and leverage a pretrained 2D CMR FM for improved accuracy. We pretrain on 160,172 multi-sequence CMR clips from 10,505 patients, including cine, LGE and T1/T2 images. Using a unified frozen-encoder architecture with gated attention aggregation, MR-JEPA is evaluated on six downstream tasks spanning ventricular function, myocardial strain, and disease classification. MR-JEPA outperforms both a natural-video foundation model (V-JEPA2, [4]) and a prior CMR-specific foundation model [22] on all regression tasks while remaining competitive for disease classification, despite using 5× fewer pretraining videos and a smaller architecture.

## 2 Related Work

### 2.1 Self-Supervised Learning for Video

Self-supervised learning (SSL) has become a powerful paradigm for learning visual representations without labels. In images, contrastive and self-distillation methods such as DINO [20, 23] learn invariant representations from augmented views using a teacher–student framework. For video, two major SSL paradigms have emerged. Masked autoencoders, including VideoMAE [25, 26], reconstruct masked spatiotemporal patches but may emphasize low-level details over semantic content. Joint-Embedding Predictive Architectures (JEPA) instead predict latent representations. I-JEPA [3] introduced this approach for images, while V-JEPA [6] and V-JEPA 2 [4] extended it to large-scale video learning, achieving strong performance on motion understanding tasks. However, these methods rely on momentum teachers, stop-gradients, and predictor networks. LeJEPA [5] recently removed these components, using a single shared encoder and Sketched Isotropic Gaussian Regularization (SIGReg) to prevent collapse, resulting in a simpler and theoretically grounded objective.

### 2.2 Foundation Models for Cardiac MRI

Several foundation models have been developed for cardiac MRI. Jacob et al. [14] introduced a DINO-based model trained on multi-sequence CMR, demonstrating benefits across segmentation and classification. However, it operates on 2D images and do not capture temporal dynamics. CineMA [12] used masked autoencoder pretraining on 15 million cine images, supporting segmentation, EF estimation, and disease detection. For video-based representation learning, Shad et al. [22] proposed a multimodal foundation model using contrastive learning over cine MRI & clinical reports. However, multimodal approaches require paired report-image data, which is often unavailable and limits scalability.

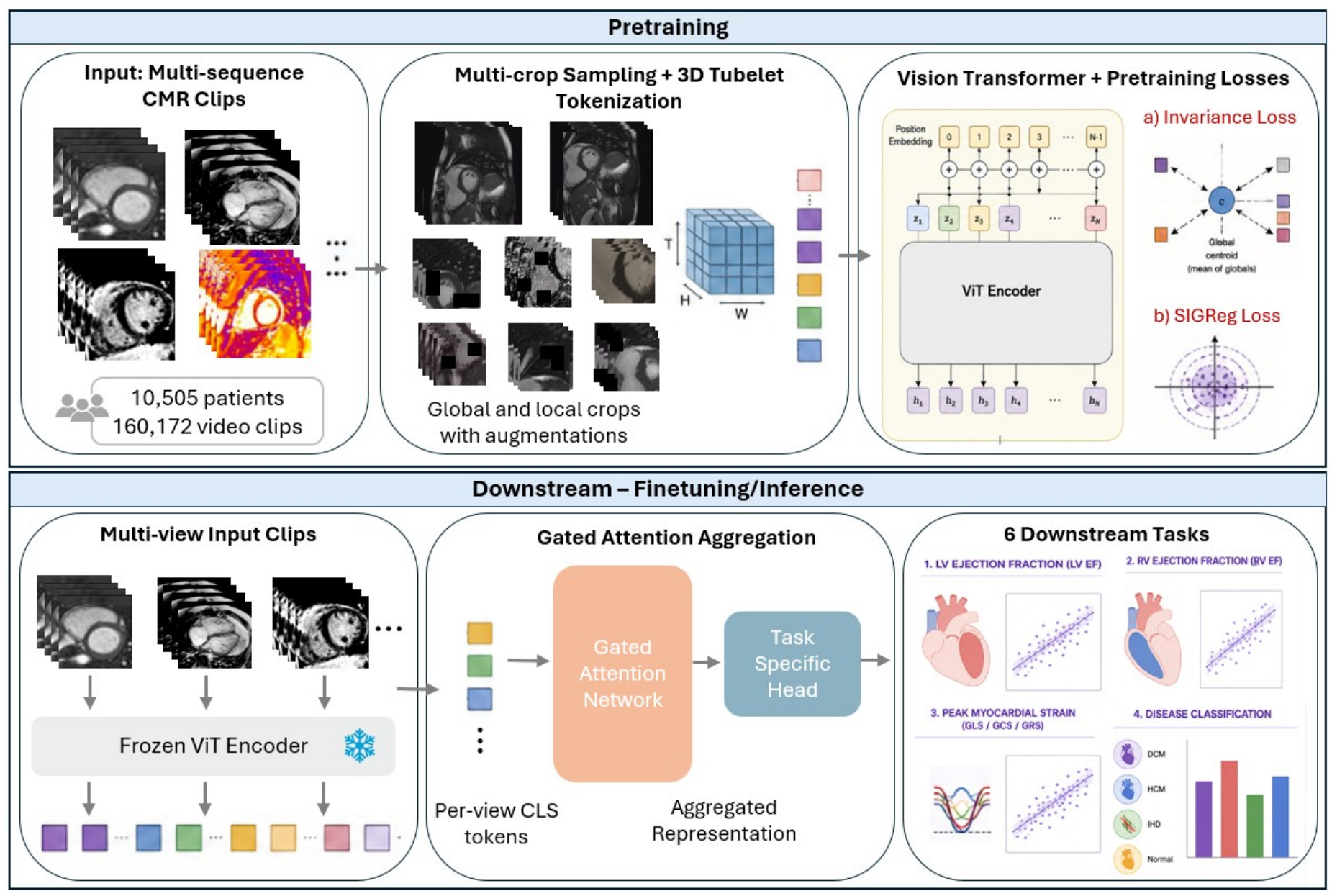

**Fig. 1.** Summary of the method - pretraining and finetuning for downstream tasks

## 3 Method

### 3.1 Pretraining

MR-JEPA extends LeJEPA [5], a joint-embedding self-supervised learning framework, from natural images to CMR video. Unlike teacher-student approaches [3, 4], LeJEPA trains a single shared encoder to learn invariant representations across multiple augmented views. Given projected embeddings $\{\mathbf{z}_v\}_{v=1}^{V}$ of all $V$ views, where the first $V_g$ correspond to global crops, the objective is

$$\mathcal{L} = (1-\lambda)\,\mathcal{L}_{\text{inv}} + \lambda\,\mathcal{L}_{\text{SIGReg}}, \tag{1}$$

with an invariance loss

$$\mathcal{L}_{\text{inv}} = \frac{1}{V} \sum_{v=1}^{V} \left\| \mathbf{z}_v - \bar{\mathbf{z}} \right\|^2, \quad \bar{\mathbf{z}} = \frac{1}{V_g} \sum_{v=1}^{V_g} \mathbf{z}_v, \tag{2}$$

and $\mathcal{L}_{\text{SIGReg}}$ regularization term [5] that prevents representation collapse by encouraging an isotropic embedding distribution without negative pairs, momentum encoders, or predictor networks.

To adapt LeJEPA to CMR video, we introduce three modifications. First, the original 2D patch embedding is replaced with a 3D tubelet embedding that jointly encodes spatial and temporal information [4, 2]. Second, we apply spatiotemporal masking to local crops, randomly masking subsets of tubelet tokens to encourage fine-grained feature learning from incomplete inputs [4]. Third, the encoder is initialized from a 2D CMR foundation model pretrained using the DINOv3 [23] objective (similar to [14]); 2D patch embeddings are inflated along the temporal dimension [8], while positional embeddings are replaced with sinusoidal spatiotemporal encodings [25].

The encoder is a Vision Transformer [11] adapted for video input via a 3D tubelet patch embedding. Input clips of consecutive grayscale CMR frames are divided into spatiotemporal tubelets, each spanning multiple frames and a spatial patch, yielding a sequence of tokens that jointly encode spatial and temporal information. A prepended classification `[CLS]` token aggregates global information, and sinusoidal positional encodings are applied separately along the spatial and temporal dimensions. The `[CLS]` output is passed through an MLP projector for the pretraining loss. MR-JEPA is pretrained on diverse CMR sequences, including cine, LGE, and T1/T2 mapping data. By jointly modeling temporal videos and spatial image stacks, the model learns a unified representation space capturing both cardiac dynamics and anatomical structure.

### 3.2 Downstream Tasks

We evaluate the frozen MR-JEPA encoder on six downstream tasks spanning cardiac function assessment and disease classification.

**Cardiac Function Regression.** Five regression tasks are considered: (1) left ventricular ejection fraction (LV EF), estimated from five cine views (2CH, 4CH, and basal, mid, and apical SAX); (2) right ventricular ejection fraction (RV EF), from 4CH and mid-SAX views; and three strain measures: (3) global longitudinal strain (GLS) from 2CH and 4CH views, and (4–5) global circumferential strain (GCS) and global radial strain (GRS) from 2CH, 4CH, and mid-SAX views.

**Cardiac Disease Classification.** The sixth task classifies patients into four categories: normal, dilated cardiomyopathy (DCM), hypertrophic cardiomyopathy (HCM), and ischemic heart disease (IHD). Inputs comprise three cine views (2CH, 4CH, mid-SAX) and a single-shot LGE (PSIR) view.

For all tasks, the frozen encoder extracts a `[CLS]` embedding from each view. View-specific embeddings are projected and aggregated using gated attention [13], enabling task-dependent weighting of complementary views. A task-specific linear head then produces either a scalar regression output or class probabilities.

## 4 Data and Experiments

### 4.1 Data

**Pretraining data.** MR-JEPA is pretrained on CMR data from two clinical centers (Center 1 and Center 2), comprising 10,505 unique patients and 160,172 video clips. The data was acquired between 2013 and 2022 on 1.5T and 3T Siemens scanners (MAGNETOM Aera, Avanto, and Skyra). Imaging protocols comprised cine bSSFP acquisitions in standard short- and long-axis views, LGE imaging using T1-weighted inversion-recovery sequences following gadolinium administration. Additional sequences included T1- and T2-weighted imaging, T2 mapping, T2* imaging etc. in Center 1, but not in Center 2. The pretraining dataset consisted predominantly of cine acquisitions (77%), followed by LGE (23%) and a small proportion of mapping data (0.4%), reflecting the distribution of routine clinical CMR examinations.

**Downstream data**

- *LV EF*: Kaggle Second Annual Data Science Bowl [19] (697 patients, USA), with expert-annotated LV volumes; split: 487/196/427. Inputs to the model consisted of 5 cine views (each view is one video): 2CH, 4CH, and 3 slices from SAX at base, mid and apex.
- *RV EF, GLS, GCS, GRS*: Center 1 (994/112/273 patients); Reference values for the measurements were obtained using a previously validated CMR analysis pipeline [10] that performed cardiac chamber segmentations on SAX and LAX views, with testing dice of LV 96.2%, myocardium 89.4%, RV 92.9%, LA 92.8%, RA 94.7% on a test set consisting of more than 1700 patients [10]. The derived measurements, including RV EF and strains, were further validated in [1, 7, 21, 15]. The input views for each individual task were chosen to be appropriate to the particular task. GCS and GRS used three input cine views: 2CH, 4CH and one SAX slice at mid-stack. GLS used 2CH and 4CH cine videos. RV EF used two views: 4CH and one SAX slice at mid-stack.
- *Disease detection*: Center 1 (942/108/256 patients). Patients were classified as normal, ischemic heart disease (IHD), hypertrophic cardiomyopathy (HCM), or dilated cardiomyopathy (DCM) by two experienced cardiologists. Class distribution was 42% normal, 33% IHD, 13% HCM, and 12% DCM. The input views consisted of three cine videos (4CH, 2CH, SAX mid-slice) and one LGE stack (singleshot stack covering the whole heart).

**Data for Model Analyses.** We study the effect of various design choices using two auxillary tasks (Refer to Section: 4.3 for more details):

*Cine view classification*: six-class classification (SAX, 2CH, 3CH, 4CH, aorta, other) on 471 cine sequences from 133 patients (Center 3), split 318/75/78.

*LGE scar detection*: binary scar classification on the EMIDEC dataset [17] (100 patients; 67% scar-positive), split 70/15/15.

### 4.2 Implementation

**Pretraining setup.** The encoder (ViT-S/16, 27.0M parameters) is trained for 300 epochs using AdamW [18] (lr $5\times10^{-4}$, weight decay 0.05, 5,000-iteration warmup, batch size 64×4 GPUs, bfloat16). Eight input frames are used. Frames are randomly sampled during training for augmentation and uniformly sampled across the full sequence during inference. Two global crops (224×224) and six local crops (96×96) are generated per sample ($V_g$=2, $V$=8). The loss weighting is $\lambda$=0.05. Training proceeds in two stages: first at half spatial resolution, then at full 224×224, loading all weights except positional embeddings (re-initialized for the new grid).

**Downstream training.** For all methods, the encoder is frozen; only the downstream head is trained (Fig. 1). Each view's `[CLS]` embedding (384-dim) is linearly projected to 512 dimensions. The gated attention mechanism uses two parallel pathways (Tanh and Sigmoid) whose element-wise product yields attention logits; softmax-normalized attention weights produce a single study-level representation. For regression, a linear head maps to a scalar and is trained using Huber loss. For classification, a linear head maps to class logits and is trained using cross-entropy with class-balanced sampling. All tasks use AdamW (lr $10^{-4}$), 300 epochs, encoder frozen.

### 4.3 Experiments

**Baselines.** The baselines use the same frozen-encoder, gated-attention downstream architecture. All compared methods are finetuned on the downstream tasks (identical data, splits) for fair comparison.

a) Shad et al. [22]: Multiscale ViT (MViT, 36.3M params) pretrained with video-text contrastive learning on 293K CMR videos from 17K patients (three US centers). Pretrained projection head is frozen and uses 16 input frames. The model was pretrained on SAX, 2CH, 3CH and 4CH by the authors.

b)V-JEPA2 [4]: ViT-B/16 pretrained on natural video using the JEPA framework, which forms a strong natural-domain baseline without medical knowledge.

**Metrics.** Regression: MAE, Pearson correlation ($r$), Bland–Altman bias and limits of agreement. Classification: macro AUC, weighted AUC, per-class one-vs-rest AUC.

**Model Analyses** We validate three design choices by comparing pretraining variants (all other settings fixed): (a) no encoder initialization from 2D CMR FM (random init); (b) no masking augmentation; (c) 16 input frames instead of 8. Each variant is evaluated via training loss and two linear probing tasks:

**Table 1.** Regression performance on the test set. $r$: Pearson correlation coefficient. Best results are shown in **bold** for MAE and $r$. Bias and limits of agreement (LoA) are reported jointly as Bland–Altman statistics. V-JEPA2 [4], Shad et al [22], MR-JEPA (ours).

| Task | Method | MAE↓ | $r$↑ | Bias (LoA) |
|---|---|---|---|---|
| LV EF ($n = 427$) | V-JEPA2 | 6.70 | 0.189 | 0.92 $[-17.6, 19.4]$ |
| | Shad et al. | 6.07 | 0.614 | 3.40 $[-11.5, 18.3]$ |
| | MR-JEPA | **4.79** | **0.764** | 0.22 $[-12.1, 12.5]$ |
| RV EF ($n = 273$) | V-JEPA2 | 9.65 | 0.323 | $-0.22$ $[-26.3, 25.8]$ |
| | Shad et al. | 9.35 | 0.410 | 0.83 $[-23.5, 25.2]$ |
| | MR-JEPA | **8.67** | **0.506** | 1.62 $[-21.5, 24.7]$ |
| GLS ($n = 273$) | V-JEPA2 | 2.38 | 0.648 | $-0.64$ $[-7.1, 5.9]$ |
| | Shad et al. | 2.90 | 0.713 | $-2.35$ $[-8.3, 3.6]$ |
| | MR-JEPA | **1.87** | **0.805** | $-0.49$ $[-5.5, 4.6]$ |
| GCS ($n = 273$) | V-JEPA2 | 3.03 | 0.564 | $-0.71$ $[-9.3, 7.9]$ |
| | Shad et al. | 2.80 | 0.677 | $-1.80$ $[-9.5, 5.9]$ |
| | MR-JEPA | **2.39** | **0.726** | $-0.46$ $[-7.6, 6.7]$ |
| GRS ($n = 273$) | V-JEPA2 | 7.45 | 0.598 | $-0.44$ $[-19.3, 18.4]$ |
| | Shad et al. | 7.26 | 0.717 | 4.16 $[-12.2, 20.5]$ |
| | MR-JEPA | **5.33** | **0.802** | $-0.97$ $[-15.0, 13.1]$ |

cine view classification representing temporal sequences, and LGE scar detection representing spatial sequences. (a) uses cine videos as inputs while (b) uses the stack of 2D LGE slices covering the spatial extent of the heart as a video input. Unlike the downstream evaluations, the model analysis uses a frozen encoder with a linear classifier to isolate representation quality rather than maximize downstream performance. Tasks from multiple CMR sequences are included to assess generalization beyond cine imaging.

## 5 Results

### 5.1 Cardiac Function Regression

Table 1 summarizes the regression results across all five cardiac function tasks. MR-JEPA achieves the best MAE and Pearson correlation on every task.

**LV Ejection Fraction.** On the Kaggle dataset ($n$=427), MR-JEPA achieves an MAE of 4.79% with a Pearson correlation of 0.764, substantially outperforming Shad et al. (MAE 6.07, $r$=0.614) and V-JEPA2 (MAE 6.70, $r$=0.189). Bland–Altman analysis shows MR-JEPA has minimal systematic bias (0.22%) and symmetric limits of agreement ($[-12.1, 12.5]$), whereas Shad et al. exhibits a systematic overestimation of 3.40%.

**RV Ejection Fraction.** RV EF is inherently more challenging due to the complex right ventricular geometry. MR-JEPA achieves the best MAE (8.67%) and correlation ($r$=0.506), improving over Shad et al. by 0.68 percentage points. All methods show wider limits of agreement compared to LV EF, reflecting the greater measurement difficulty.

**Table 2.** Disease classification performance on the held-out test set ($n = 256$), reported as macro AUC, weighted AUC, and one-vs-rest (OvR) AUC for each disease class. Best results are shown in **bold**. V-JEPA2 [4], Shad et al [22], MR-JEPA (ours).

| Metric | V-JEPA2 | Shad et al. | MR-JEPA |
|---|---|---|---|
| Macro AUC | 0.741 | **0.882** | 0.868 |
| Weighted AUC | 0.743 | **0.864** | 0.854 |
| OvR AUC – Normal | 0.773 | 0.888 | **0.889** |
| OvR AUC – DCM | 0.777 | **0.905** | 0.890 |
| OvR AUC – HCM | 0.708 | **0.950** | 0.922 |
| OvR AUC – IHD | 0.705 | **0.785** | 0.771 |

**Myocardial Strain.** The largest improvements were observed for myocardial strain estimation. MR-JEPA achieved the best performance on GLS, GCS, and GRS, with MAEs of 1.87, 2.39, and 5.33, respectively. Relative to the strongest baseline, this corresponds to MAE reductions of 15–27% and higher correlations across all strain tasks ($r = 0.726$–$0.805$). MR-JEPA also maintained low systematic bias ($|\text{bias}| \leq 0.97$).

### 5.2 Cardiac Disease Detection

Table 2 presents the disease classification results. Unlike the regression tasks, Shad et al. achieves the highest macro AUC (0.882 vs. 0.868 for MR-JEPA), although MR-JEPA substantially outperforms V-JEPA2 (0.741). MR-JEPA achieves the highest one-vs-rest AUC for the normal class (0.889) and remains within 0.01–0.03 AUC of Shad et al. across all disease classes. These results suggest that MR-JEPA learns representations that transfer effectively to disease classification despite being optimized using a fully self-supervised objective. The stronger performance of Shad et al. on disease classification, despite not being pretrained on LGE, may be attributable to several differences in the pretraining setup, including a larger pretraining dataset (293K vs. 160K clips), longer input sequences (16 vs. 8 frames), and the use of video–text contrastive learning with clinical report supervision. Nevertheless, MR-JEPA remains competitive despite using fewer pretraining videos, a smaller model (27M vs. 36.3M parameters), and no text supervision.

### 5.3 Model Analysis

Table 3 evaluates three design choices of MR-JEPA. Although pretraining losses were similar across configurations (0.207–0.221), downstream performance differed markedly. Removing 2D-CMR foundation model initialization or spatiotemporal masking substantially reduced transferability, with scar detection dropping from 0.850 to chance level (0.500) in both cases. Increasing clip length from 8 to 16 frames reduced optimization efficiency (103 vs. 190 epochs) and lowered view classification accuracy (0.841 vs. 0.932). Overall, 2D-CMR initialization, spatiotemporal masking, and 8-frame clips yielded the strongest representations.

**Table 3.** Hyperparameter study. Linear probing accuracy for cine view classification and LGE scar detection under different pretraining configurations.

| Configuration | Pretrain Loss | Epochs | View Cls. Acc.↑ | Scar Det. Acc.↑ |
|---|---|---|---|---|
| No DINOv3 init | 0.208 | 207 | 0.890 | 0.500 |
| No masking | 0.207 | 223 | 0.908 | 0.500 |
| 16 frames | 0.221 | 103 | 0.841 | 0.750 |
| **Ours (full)** | 0.215 | 190 | **0.932** | **0.850** |

## 6 Conclusions

We present MR-JEPA, a video foundation model for cardiac MRI that extends LeJEPA to 3D spatiotemporal inputs. Using a simple joint-embedding objective with spatiotemporal masking and initialization from a 2D CMR foundation model, MR-JEPA learns representations from 160K multi-sequence CMR clips across 10,505 patients without annotations. With a frozen encoder and gated attention aggregation, it achieves state-of-the-art performance on five cardiac function regression tasks (LV EF, RV EF, GLS, GCS, GRS), outperforming domain-specific and natural video baselines. Compared with prior CMR video foundation models trained exclusively on cine imaging, pretraining on multiple CMR sequences may improve representation learning across heterogeneous image contrasts. On disease detection, it remains competitive despite using less data and a simpler, unsupervised training objective. Ablations confirm the importance of encoder initialization, spatiotemporal masking, and 8-frame inputs, while embedding analysis shows that the model naturally separates temporal and spatial sequences and generalizes to unseen spatial cine volumes.

*Limitations:* Pretraining was performed using single-vendor CMR data. Although the LV EF task was evaluated on an independent multi-center dataset, the remaining downstream tasks were evaluated on data from one pretraining center; however, no labels were used during pretraining. Ground truth for RV EF and strain estimation was obtained from a previously validated deep learning model rather than manual expert annotations. Finally, the pretraining dataset was heavily dominated by cine acquisitions, with only a small proportion of mapping data, that too from a single center. Future work should collect more mapping data and evaluate on mapping-specific downstream tasks (e.g., fibrosis/edema quantification). Finally, we compared MR-JEPA primarily against existing foundation models under a unified downstream evaluation protocol. Comparison with task-specific state-of-the-art models would provide additional clinical context and is left for future work.

**Disclaimer**: The concepts and information presented in this paper are based on research results that are not commercially available. Future commercial availability cannot be guaranteed. **Disclosure of Interests**: AJ, PS and DC are employees of Siemens Healthineers. The other authors declare that they have no competing interests.